\documentclass{article}

\usepackage[preprint]{corl_2024} % Uncomment for pre-prints (e.g., arxiv); This is like ``final'', but will remove the CORL footnote.

\usepackage{bm}
\usepackage{amsmath}
\usepackage{graphicx}
\usepackage{graphicx}
\usepackage{subcaption}

\usepackage{xspace}
\usepackage{amssymb}
\usepackage{multirow}
\usepackage[acronym]{glossaries}
\usepackage[inline]{enumitem}
\usepackage{caption}
\usepackage{booktabs}
\usepackage{setspace}
\usepackage{wrapfig}
\usepackage{afterpage} % make it easy to insert new page after a figure
\usepackage{setspace}
\usepackage{float}     % For [H] specifie

\title{BMP: Bridging the Gap between B-Spline and Movement Primitives}

\author{
  Weiran Liao, \ Ge Li, \ Hongyi Zhou, \ Rudolf Lioutikov, \ Gerhard Neumann\\
  Karlsruhe Institute of Technology\\
  \texttt{ge.li@kit.edu}\\
  }

\begin{document}
\maketitle

%===============================================================================

\begin{abstract}
    % Movement Primitives (MPs) are a fundamental concept for compact movement modeling, which can roughly categorized into dynamic system-based MPs and Probabilistic-based MPs.
    % The Dynamic Movement Primitives (DMPs) can generate a trajectory smoothly starting from a given initial condition and converging to a goal but do not support probabilistic modeling. In contrast, the Probabilistic Movement Primitives (ProMPs) provide probabilistic modeling but can not ensure a precise smooth start. In this paper, we adopt the ProMP framework and use the B-spline as MPs, which keeps probabilistic modeling while satisfying any boundary conditions. We demonstrate using B-spline in a probabilistic manner, showing it is a unified MP formulation for imitation learning and reinforcement learning.
    This work introduces \textbf{B-spline Movement Primitives (BMPs)}, a new Movement Primitive (MP) variant that leverages B-splines for motion representation. B-splines are a well-known concept in motion planning due to their ability to generate complex, smooth trajectories with only a few control points while satisfying boundary conditions, i.e., passing through a specified desired position with desired velocity. However, current usages of B-splines tend to ignore the higher-order statistics in trajectory distributions, which limits their usage in imitation learning (IL) and reinforcement learning (RL), where modeling trajectory distribution is essential. In contrast, MPs are commonly used in IL and RL for their capacity to capture trajectory likelihoods and correlations. However, MPs are constrained by their abilities to satisfy boundary conditions and usually need extra terms in learning objectives to satisfy velocity constraints. By reformulating B-splines as MPs, represented through basis functions and weight parameters, BMPs combine the strengths of both approaches, allowing B-splines to capture higher-order statistics while retaining their ability to satisfy boundary conditions. Empirical results in IL and RL demonstrate that BMPs broaden the applicability of B-splines in robot learning and offer greater expressiveness compared to existing MP variants.

\end{abstract}


%===============================================================================

\section{Introduction}

\textbf{Movement Primitives (MPs)}\citep{Ijspeert_2013,NIPS2013_e53a0a29,zhou2019learning,li_prodmp_2023} are a well-established concept in robot learning, offering a compact parameterization of movements that can be efficiently reused, adapted and combined to synthesize complex behavior. They can be categorized into dynamic system-based MPs and probabilistic-based MPs \cite{li_prodmp_2023}. 
Dynamic system-based MPs, model movements as dynamic systems with an emphasis on ensuring global stability. The first of this kind is Dynamic Movement Primitives (DMP) \cite{Ijspeert_2013}. DMP employs a second-order dynamic system that guarantees trajectory convergence to a goal attractor. While DMPs are effective for point-to-point tasks, they are less suited for more general motion tasks that require passing through several via points. Additionally, DMPs lack probabilistic modeling capabilities due to their dynamic system formulation, limiting their applicability in tasks that require trajectory distribution modeling, such as imitation learning (IL) and reinforcement learning (RL).
On the other hand, probabilistic-based MPs model movements directly as trajectory distributions, allowing for flexibility in maintaining and adapting these distributions. A key example is Probabilistic Movement Primitives (ProMPs)\cite{NIPS2013_e53a0a29}, which represent trajectories as a linear combination of basis functions with weights. ProMPs use a linear Gaussian model, enabling probabilistic inference to handle complex interactions and adapt trajectories based on context. While ProMPs excel in modeling uncertainty and adapting to new situations, they struggle with satisfying boundary conditions like initial states, particularly in reinforcement learning tasks where robots are reset to different initial states at the beginning of each episode.

In the realm of trajectory planning, B-splines are widely applied due to their
% compact representation, 
minimal parameter requirements and local support of basis functions, which benefit optimization processes. In addition, B-splines can satisfy any boundary conditions, i.e., passing through specified positions with desired velocity, making them ideal for planning and replanning smooth trajectories. However, current usages of B-splines focusing on model single trajectory, ignoring the trajectory distributions \cite{usenko_real-time_2017,zhou_robust_2019, kicki_fast_2023, kicki2024bridginggaplearningtoplanmotion}, limiting their usages in IL and RL where trajectories likelihood and temporal correlations are desired \cite{li_prodmp_2023, Li_tce_2024}. 

In this paper, we propose BMPs, a novel approach that integrates B-splines as probabilistic-based Movement Primitives. This unified framework is suitable for both imitation learning (IL) and episodic reinforcement learning (ERL) \cite{Li_tce_2024, Otto_bbrl_2022, otto2023mp3, li2024top}, retaining the strengths of MPs—such as probabilistic representation—while also ensuring the satisfaction of boundary conditions. Through three experiments, we demonstrate the effectiveness of BMPs in both IL and ERL settings, showcasing their broader applicability and expressiveness in robot learning.

% Moreover, since B-splines also represent trajectories as a linear combination of basis functions, they can naturally adopt the probabilistic modeling framework used in ProMPs.

    % Recently, a framework called Probabilistic Dynamic Movement Primitives (ProDMPs) \citep{li_prodmp_2023} has been developed, which provides a unified framework that retains the benefits of both ProMP and DMP, combining the flexibility of probabilistic modeling with the ability to satisfying an initial condition.

    % it should be able to scale in time, allowing the movement to slow down or speed
    % up. It should also adapt to different starting states and goals, providing flexibility.
    % Additionally, MPs should be composable, meaning they can be combined sequentially
    % or added together to create new and more complex movements. 
	
    % Submission to CoRL 2024 will be entirely electronic, via a web site (not email). Information about the submission process and \LaTeX{} templates are available on the conference web site at \url{https://corl2024.org/}. For camera ready submission, use the \texttt{final} option for the \texttt{\textbackslash usepackage} command. 

%===============================================================================

\section{Related Works}

\textbf{Dynamic Movement Primitives (DMPs)} \citep{Ijspeert_2013} model a trajectory as second-order dynamical systems (DS) with asymptotically vanishing nonlinear forcing input term. it is designed to ensure asymptotic stability, that all executing movements converge asymptotically to the goal attractor. Once a DMP is learned, to apply it to reproduce the movement adaptively as its original design purpose, it should be used in real-time integration form with real feedback values. 
Considering perturbations during movement execution, they are well-suited for point-to-point tasks, where the primary requirement is that the movement reaches a desired goal state. Many tasks, however, require the robot to dynamically follow desired trajectories but not just to converge to a fixed point, i.e. the shape of the movement after perturbations is essential for the tasks. There are various modification of DMP for different applications summarized in \citep{saveriano_dynamic_2021}.   

In imitation learning (IL), through setting the initial condition and goal atrractor of the DS, DMP has the advantage of adapting learned movement to new start positions, velocity, and new end positions while keeping the learned movement shape. In RL or general planning tasks, DMP ensures a precise smooth start. Notably, The forcing term containing the parameters shaping the trajectory is the input to the dynamic system. This makes the learning process need to propagate through the dynamic system and, hence makes DMPs unable to capture trajectory statistics.

\textbf{Probabilistic Movement Primitives (ProMPs)}\citep{NIPS2013_e53a0a29} directly represent a trajectory by a linear combination of basis functions with weights and thus utilize a linear Gaussian model to model the trajectory distribution. A single Degree of Freedom (DoF) example can be written as
\begin{equation}
    \bm{y}_{0:T} = \bm{\Phi}_{0:T}^\top \bm{\omega} + \bm{\epsilon_y}, \label{eq:prompy}
\end{equation}
\begin{equation}
    p(\bm{y}_{0:T} | \bm{\mu}_{\bm{\omega}}, \bm{\Sigma}_{\bm{\omega}}) = \mathcal{N}(\bm{y}_{0:T}| \bm{\Phi}_{0:T}^\top \bm{\mu}_{\bm{\omega}}, \bm{\Phi}_{0:T}^\top \bm{\Sigma}_{\bm{\omega}} \bm{\Phi}_{0:T} + \bm{I} \sigma_y^2 ), \label{eq:prompd}
\end{equation}
where $\bm{y}_{0:T}$ is the desired trajectory evaluated at time steps from 0 to $T$. $\bm{\Phi}_{0:T}$ defines the $n \times T$ dimensional basis matrix consisting of evaluated values at all time steps of $n$ normalized radial basis function (RBF). $\epsilon_y \sim (0,\sigma_y)$ is a zero-mean i.i.d. Gaussian noise. 
$\bm{\omega}$ defines an $n$-dimensional parameters vector that fellows a Gaussian distribution $\mathcal{N}(\bm{\omega} | \bm{\mu}_{\bm{\omega}}, \bm{\Sigma}_{\bm{\omega}})$.
The \eqref{eq:prompd} can be extended to model multiple DoF trajectories as
% \begin{equation}
%     % \resizebox{\textwidth}{!}{$
%     p(\bm{Y}_{0:T} | \bm{\mu}_{\bm{\Omega}}, \bm{\Sigma}_{\bm{\Omega}}) = 
%     \mathcal{N} \left( 
%     \begin{bmatrix}
%         \bm{y}_{1,0:T} \\
%         \vdots \\
%         \bm{y}_{d,0:T}
%     \end{bmatrix} 
%     \Bigg| 
%     \begin{bmatrix}
%         \bm{\Phi}_{0:T}^T & \cdots & 0 \\
%         \vdots & \ddots & \vdots \\
%         0 & \cdots & \bm{\Phi}_{0:T}^T
%     \end{bmatrix} 
%     \bm{\mu}_{\bm{\Omega}},
%     \begin{bmatrix}
%         \bm{\Phi}_{0:T}^T & \cdots & 0 \\
%         \vdots & \ddots & \vdots \\
%         0 & \cdots & \bm{\Phi}_{0:T}^T
%     \end{bmatrix}
%     \bm{\Sigma_\Omega}
%     \begin{bmatrix}
%         \bm{\Phi}_{0:T}^T & \cdots & 0 \\
%         \vdots & \ddots & \vdots \\
%         0 & \cdots & \bm{\Phi}_{0:T}^T
%     \end{bmatrix}^{\top} + \bm{\Sigma}_y
%     \right)
%     % $}
%     \label{eq:promulti}
% \end{equation}

\begin{equation}
    \begin{split}
        p(\bm{Y}_{0:T} | \bm{\mu}_{\bm{\Omega}}, \bm{\Sigma}_{\bm{\Omega}}) 
        &= \mathcal{N} \left( 
        \begin{bmatrix}
            \bm{y}_{1,0:T} \\
            \vdots \\
            \bm{y}_{d,0:T}
        \end{bmatrix} 
        \Bigg| 
        \begin{bmatrix}
            \bm{\Phi}_{0:T}^T & \cdots & 0 \\
            \vdots & \ddots & \vdots \\
            0 & \cdots & \bm{\Phi}_{0:T}^T
        \end{bmatrix} 
        \bm{\mu}_{\bm{\Omega}}, \right.\\
        & \quad \left. \begin{bmatrix}
            \bm{\Phi}_{0:T}^T & \cdots & 0 \\
            \vdots & \ddots & \vdots \\
            0 & \cdots & \bm{\Phi}_{0:T}^T
        \end{bmatrix}
        \bm{\Sigma_\Omega}
        \begin{bmatrix}
            \bm{\Phi}_{0:T}^T & \cdots & 0 \\
            \vdots & \ddots & \vdots \\
            0 & \cdots & \bm{\Phi}_{0:T}^T
        \end{bmatrix}^{\top} + \bm{\Sigma}_y
    \right)
    \end{split}
    \label{eq:promulti}
\end{equation}

where $d$ indicates the DoF, $ \bm{\mu}_{\bm{\Omega}} = [\bm{\mu}_{\bm{\omega}_1}^T, \dots, \bm{\mu}_{\bm{\omega}_d}^T]^T $. $\bm{\Sigma_\Omega}$ is a $(d \times n) \times (d \times n)$ multi-DoF weights covariance matrix capturing also the correlation across DoFs. By maintaining the trajectory distributions, ProMPs can generate trajectories conditioned on observed trajectory values. However, ProMP is unable to satisfy precise boundary conditions, because the normalized RBF function never reaches exactly zero value.

\textbf{Probabilistic Dynamic Movement Primitives (ProDMPs)} \citep{li_prodmp_2023} is proposed to provide a unified framework that retains the benefits of both ProMP and DMP, combining the flexibility of probabilistic modeling with the ability to satisfy an initial condition. It solves the differential equation of DMP to construct a set of basis functions in position space, thus trajectories can be represented in a similar form to ProMP,
\begin{equation}
     y(t) = c_1 y_1(t) + c_2 y2(t) + \bm{\Phi}_{prodmp}^T(t) \bm{\omega}_g, \label{eq:prodmp}
\end{equation}
where $c_1 y_1(t) + c_2 y2(t)$ ensures satisfying initial position and velocity conditions. Although it is only used as a pure trajectory generator like ProMP in the ProDMP paper, the learned weights can still applied to a DMP in adaptive manner, since all basis functions are directly solved from the DMP formulation. However, the ProDMP can only satisfy initial conditions. In addition, the basis functions $\bm{\Phi}_{prodmp}$ exhibit substantial magnitude differences among basis, which need to be rescaled in RL practice. And the basis functions overlap with each other throughout the whole movement duration, which is undesired since changing one weight will affect the whole trajectory.

%===============================================================================
\begin{figure}[t!]
    \centering
    \begin{subfigure}[b]{0.24\textwidth}
        \centering
        \includegraphics[width=\linewidth]{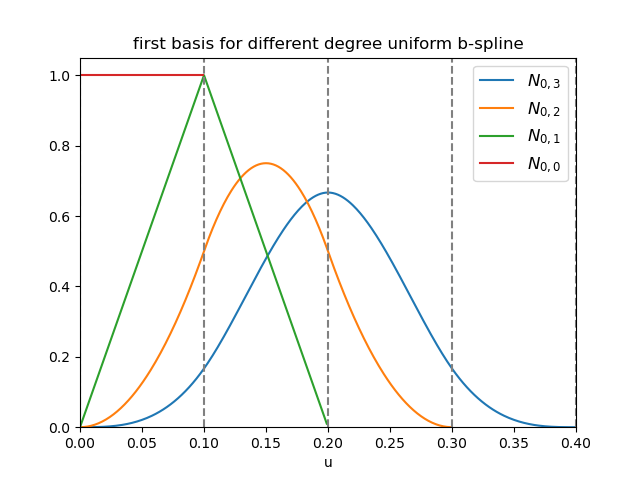}
        \caption{}
        % \caption{The first basis for B-spline of degree 0， 1， 2， 3}
        \label{fig:n00}
    \end{subfigure}
    \hfill
    \begin{subfigure}[b]{0.24\textwidth}
        \centering
        \includegraphics[width=\linewidth]{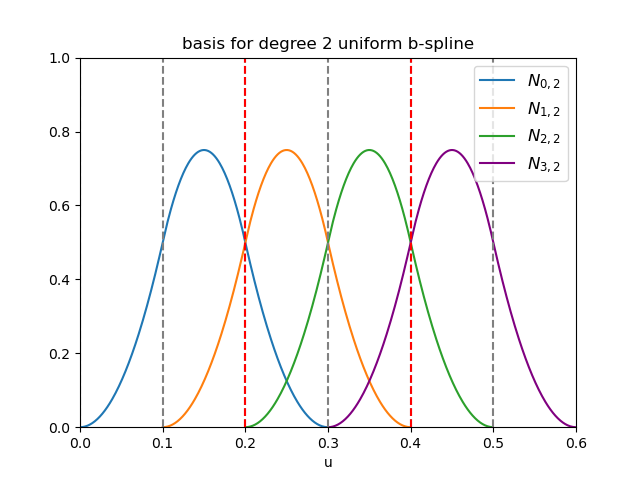}
        \caption{}
        % \caption{Basis functions for a degree 2 uniform B-spline with 4 basis defined on knot vector $[0,\ 0.1, \ 0.2, \ 0.3, \ 0.4, \ 0.5, \ 0.6]$, the valid spans for representing a trajectory are $[0.2,\ 0.4]$, where there are always 3 non-zero bases in each span}
        \label{fig:valid_b}
    \end{subfigure}
    \hfill
    \begin{subfigure}[b]{0.24\textwidth}
        \centering
        \includegraphics[width=\linewidth]{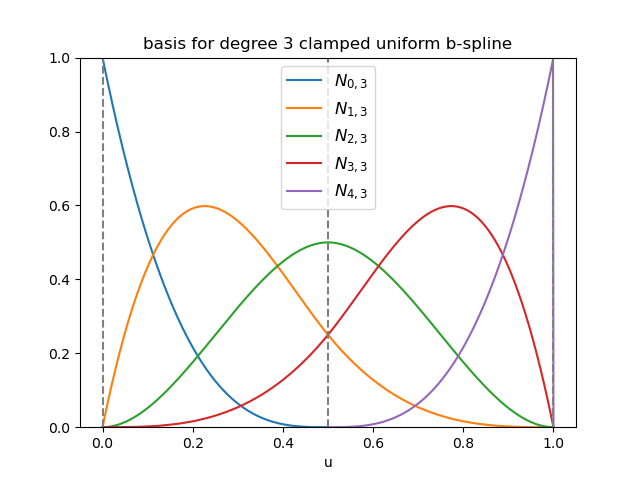}
        \caption{}
        % \caption{5 Basis functions clamped uniform B-spline of degree 3 defined on knot vector $[0,\  0,\  0,\  0,\  0.5,\  1,\  1,\  1,\  1]$}
        \label{fig:clamped}
    \end{subfigure}
    \hfill
    \begin{subfigure}[b]{0.24\textwidth}
        \centering
        \includegraphics[width=\linewidth]{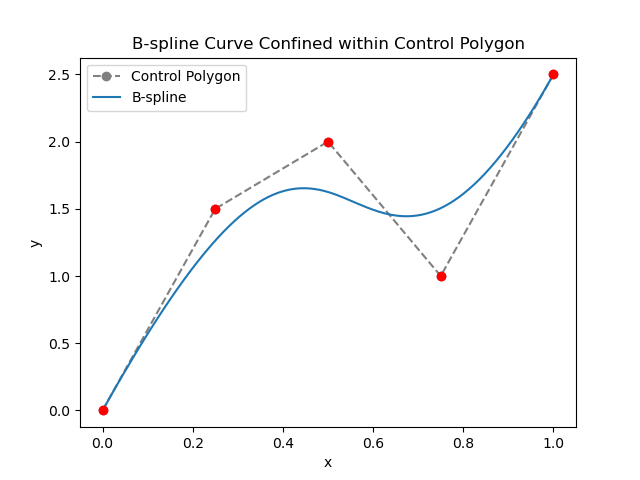}
        \caption{}
        % \caption{The Clamped B-spline trajectory starts exactly from first control point and ends at the last control point, the whole trajectory stays inside the polygon region of control points.}
        \label{fig:convex}
    \end{subfigure}
    \caption{(a) The first basis for B-spline of degree 0, 1, 2, 3. (b) Basis functions for a degree 2 uniform B-spline with 4 basis defined on knot vector $[0,\ 0.1, \ 0.2, \ 0.3, \ 0.4, \ 0.5, \ 0.6]$, the valid spans for representing a trajectory are $[0.2,\ 0.4]$, where there are always 3 non-zero bases in each span. (c) Clamped uniform B-spline of degree 3  with 5 Basis defined on knot vector $[0,\  0,\  0,\  0,\  0.5,\  1,\  1,\  1,\  1]$. (d)The Clamped B-spline trajectory starts exactly from first control point and ends at the last control point, the whole trajectory stays inside the polygon region of control points}
    % \label{fig:main_label}
\end{figure}

\section{Bridging the Gap between B-Spline and Movement Primitives}

\textbf{B-spline} \citep{prautzsch2002bezier} is a piecewise polynomial function commonly used for approximating curves and surfaces. 
% Its mathematical formulation is based on the concept of \textit{basis functions} and \textit{control points}. 
For a B-spline of degree $p$ (where $p \geq 0$), the B-spline curve $y(u)$ is expressed as a linear combination of basis functions:
\begin{equation}
    y(u) = \sum_{i=0}^{n-1} N_{i,p}(u) \  c_i, \quad 0<p<n, \quad u \in [u_0, u_m].
\end{equation}
where $c_i$ are the control points, the weights for corresponding basis functions. $N_{i,p}(u)$ are the B-spline basis functions of degree $p$, defined on the knot vector $\bm{u} = [u_0, ..., u_m]$, with $u_{i} \le u_{i+1}$.
$u$ is a generalized time variable within the interval $[u_0, u_m]$.
% The generalized time variable $u$ lies within the interval$[u_0, u_m]$, with $u_{i} \le u_{i+1}$.
% typically normalized between the first and last knot values
% In the context of B-splines, the $ \bm{c} = [c_0, ..., c_{n-1}]^\top$ are called control points.
The basis functions $N_{i,p}(u)$ are defined recursively using Cox–de Boor recursion formula \citep{prautzsch2002bezier}. For degree $p=0$, the basis functions are piecewise constant:
\begin{equation}
    N_{i,0}(u) = 
    \begin{cases} 
    1 & \text{if } u_i \leq u < u_{i+1},  \text{ if } u_i \leq u \leq u_{i+1} \text{ for } i=n-1, \\
    0 & \text{otherwise}.
    \end{cases}
    \label{eq:deboor_constant}
\end{equation}
For degree $p>0$, the recursive definition is given by:
\begin{equation}
    N_{i,p}(u) = \frac{u - u_i}{u_{i+p} - u_i} N_{i,p-1}(u) + \frac{u_{i+p+1} - u}{u_{i+p+1} - u_{i+1}} N_{i+1,p-1}(u),    
    \label{eq:deboor_recursive}
\end{equation}
For example, as shown in \ref{fig:n00}, the degree 0 basis function $N_{0,0}$ is a step function valid in the knot span $[u_0,u_1)$. 
The degree 1 basis function $N_{1,0}$ is a triangle spanning $[u_0,u_2)$, with its peak at at $u_1$. 
The degree 2 basis function $N_{2,0}$ forms a parabola over $[u_0,u_3)$.

In each valid knot span for modeling, there are $p+1$ non-zero basis functions. For a degree 2 B-spline, for instance in \ref{fig:valid_b}, the first valid span is $[u_2, u_3)$, and the last is $[u_{m-3},u_{m-2}]$.
Generally, for a degree $p$ B-spline with $n$ basis, the knot vector $\bm{u} = [u_0, ..., u_m]$ contains $m+1$ knots, satisfying the condition $m=n+p$. 
The valid spans are $[u_{p}, ..., u_{m-p}]$, and within each span $[u_i, u_{i+1})$, the curve is influenced by $p+1$ control points $c_{i-p}, ..., c_{i}$.
An important property of the B-spline is the local support of the basis functions—each basis function only affects a limited portion of the curve, providing local control. 
Additionally, due to the recursive definition, the derivative of a B-spline curve of degree $p$ with $n$ control point is a B-spline of degree $p-1$ with $n-1$ control points, allowing easy derivation of the control points and basis functions of the derivative. Finally, B-splines possess the convex hull property, meaning the curve is contained within the convex hull of its control points shown in \ref{fig:convex}. This property is frequently used to efficiently impose bound constraints.

We adopt the clamped uniform B-spline as MP. Uniform means all knots in the valid spans are equidistantly located.  
Clamped means that the curve passes through the first and last control points as shown in \ref{fig:convex}, which is utilized to satisfy boundary conditions. This is achieved by repeating both the first and last knot $p+1$ times, as demonstrated in \ref{fig:clamped}, ensuring that the first basis function evaluates to 1 at time point $u=0$ while the other basis functions are evaluated to 0, and the same applies reversely at the end time point.
% This is achieved by repeating both the first and last knot $p+1$ times, also known as the multiplicity of the starting and ending knots, ensuring that the first basis function evaluates to 1 at the starting knot $u_{p}$, while the other $p-1$ basis functions evaluate to 0. The same applies at the end of the curve, where the final basis function evaluates to 1 at the last knot. For example, the basis functions of a clamped uniform b-spline of degree 3 with 5 basis are given in \Cref{fig:cbspb}.
% and the trajectory start exactly from the first control point and ends at the last control point as shown in \cref{fig:cbsp_ch}.

\textbf{BMP: Using B-spline to extend ProMP}. For a single DoF trajectory, the B-spline can be write in a similar form to ProMP as
% \begin{equation}
%     y(u) = \sum_{i=0}^{n-1} N_{i,p}(u) \  \omega_i, \quad 0<p<n, \quad u \in [0, 1].
% \end{equation}
\begin{equation}
    \bm{y}_{0,...T} = [\bm{N}_{0,p}(u_{0,...,T}) \ ... \  \bm{N}_{n-1,p}(u_{0,...,T})] \begin{bmatrix} c_0 \\ \vdots \\ c_{n-1} \end{bmatrix} = \bm{\Phi}^{T}(u_{0,...,1}) \ \bm{c} \label{eq:bmp} 
\end{equation}
where $y_{0,...T}$ are the trajectory values in the time steps $[0,...,T]$, the $u_{0,...,T}$ are linear phases values with $u(t)=\frac{t-t_0}{T}$ ranging from 0 to 1. The B-spline basis function $N_{i,p}(u)$ is defined on a knot vector $\bm{u} = [u_0, u_1,...,u_p, ..., u_{m-p},...,u_{m-1}, u_m]$, where $u_0, u_1,...,u_p = 0$, $u_{m-p},...,u_{m-1}, u_m = 1$, and $u_{i+1}-u_i=\Delta=\frac{1}{m-2p}$ for $i \in [p,...,m-p-1]$. 
% It is worth mentioning, to construct the basis function, we adopt a modification in the original Cox–de Boor formula \Cref{eq:deboor}, where for $i=n-1$
% \begin{equation}
%     N_{n-1,0}(u) = 
% \begin{cases} 
% 1 & \text{if } u_{n-1} \leq u \leq u_{n}, \\
% 0 & \text{otherwise}
% \end{cases}.
% \end{equation}
% This ensures the last basis function $N_{n-1,p}$ evaluated at the right bound $u=1$ equals 1, but not undefined in the original formula, which can cause a sudden movement jump to 0 at the end.

Utilizing the derivative property of the B-spline, we have the trajectory velocity with respect to phase in the form of
% \begin{equation}
%     \dot{\bm{y}}_{0,...T} = [\bm{N}_{1,p-1}(u_{0,...,T}) \ ... \  \bm{N}_{n-1,p-1}(u_{0,...,T})] \begin{bmatrix} c_0^{(1)} \\ \vdots \\ c_{n-2}^{(1)} \end{bmatrix} = \dot{\bm{\Phi}}^{T}(u_{0,...,1}) \ \bm{c}^{(1)},
% \end{equation}
\begin{equation}
    \dot{\bm{y}}_{0,...T} = [\bm{N}_{1,p-1}(u_{0,...,T}) \ ... \  \bm{N}_{n-1,p-1}(u_{0,...,T})] \begin{bmatrix} c_0^{(1)} \\ \vdots \\ c_{n-2}^{(1)} \end{bmatrix} = \dot{\bm{\Phi}}^{T}(u_{0,...,1}) \ \bm{c}^{(1)},
\end{equation}
where the velocity control point is computed through
\begin{equation}
    {c}_i^{(1)} = \frac{p}{\Delta} ({c}_{i+1} - {c}_i). \label{eq:49}
\end{equation}
% Recursively, we can also get the trajectory acceleration in the form of
% \begin{equation}
%     {y}^{(2)}_{0,...T} = [N_{2,p-2}(u_{0,...,T}) \ ... \  N_{n-1,p-2}(u_{0,...,T})] \begin{bmatrix} c_0^{(2)} \\ \vdots \\ c_{n-3}^{(2)} \end{bmatrix} = \bm{\Phi}^{(2)}^{T}(u_{0,...,1}) \ \bm{c}^{(2)},
% \end{equation}
% where the acceleration control point is computed as
% \begin{equation}
%     {c}_i^{(2)} = \frac{p}{\Delta} ({c}_{i+1}^{(1)} - {c}_i^{(1)}).
% \end{equation}
This can be recursively applied to obtain the acceleration profile.
The trajectory starts exactly from the first control point $c_0$ and ends at the last control point $n_{n-1}$. The velocity profile starts exactly from the first velocity control point $c^{1}_0$ and ends at the last velocity control point $c^{1}_{n-2}$. In the above derivation, the velocity are computed with respect to the phase but not the time. It needs to be divided by the duration $T$ to get real velocity.

Depending on the initial conditions and end conditions we are about to impose, we can prescribe values to the first control points and the last control points. To impose initial position $y_0$ and initial velocity $\dot{y}_0$, we solve the linear equations
\begin{equation} 
\begin{cases} 
c_0= y_0 \\
c^{(1)}_0= \frac{p}{\Delta} (c_1 - c_0) = \dot{y}_0
\end{cases} \ 
\Rightarrow \ 
\begin{cases} 
c_0 = y_0 \\
c_1 = \frac{\Delta}{p}\dot{y}_{0} +  c_0
\end{cases},
\end{equation}
and set the first two control points respectively.
The same applies reversely for imposing end position or end position plus end velocity.
Sometimes, only the end velocity but no end position is to be prescribed, for example, when a zero end velocity is desired but the position should be learned. It can be solved by
\begin{equation}
    c^{(1)}_{n-2} = \frac{p}{\Delta} (c_{n-1} - c_{n-2}) = \dot{y}_{e} \ \Rightarrow \ c_{n-2} = -\frac{\Delta}{p}\dot{y}_{e} +  c_{n-1}.
\end{equation}
where $\dot{y}_{e}$ is the desired end velocity. This means the second last control point $c_{n-2}$ is set as a pre-given constant $-\frac{\Delta}{p}\dot{y}_{e}$ plus the later learned last control point value $c_{n-1}$. It is obvious that the to-be-learned weights are these control points in between and maybe also the last control point. 
% The single DoF trajectory distribution takes the same form as \eqref{eq:prompd}
% , but only considers the to-be-learned weights.

Due to the linear combination representation, the linear Gaussian modeling in ProMP can be applied to BMP. For instance, in the case of given initial position and velocity conditions, end position and velocity conditions, 
the single DoF trajectory distribution takes a similar form to \eqref{eq:prompd}:
\begin{equation}
    p(\bm{y}_{0:T} | \bm{\mu}_{\bm{c}_{2:n-3}}, \bm{\Sigma}_{\bm{c}_{2:n-3}}) = \mathcal{N}(\bm{y}_{0:T}| \bm{\Phi}_{0:T,[2:n-3]}^\top \bm{\mu}_{\bm{c}_{2:n-3}} + \bm{d}, \bm{\Phi}_{0:T,[2:n-3]}^\top \bm{\Sigma}_{\bm{c}_{2:n-3}} \bm{\Phi}_{0:T,[2:n-3]} + \bm{I} \sigma_y^2 ),
\end{equation}
where $\bm{d}$ is the deterministic part of the trajectory, defined by boundary conditions.
This can be extended to the multi-DoF case to model the uncertainty and correlation across both dimension and time, following the same way as in \eqref{eq:promulti}.
% To model the uncertainty and correlation cross dimension and time in a multi-DoF case, 
% only these basis functions with to-be-learned control point values associated to them, are sliced out from the basis matrix $\bm{\Phi}(u_{0,...,1})$ in \eqref{eq:bmp} and extended to multi-DoF to construct a reduced diagonal block matrix as in ProMP \eqref{eq:promulti}.

In the practical implementation, the trajectory is composed of a constant line with initial position value $y_{0}$ and the linear combination of basis functions with control points, i.d. the B-spline only learns the residual based on the initial position and the initial position feed to B-spline is always 0. This makes the variance of the to-be-learned control points smaller, which significantly improves the learning performance compared to simply using the B-spline. Additionally, the constant line can be also replaced by a linear trajectory pointing from the initial position to the goal position.

%===============================================================================
\section{Experiments}
\subsection{B-spline versus MPs}
\begin{figure}[ht!]
    \centering
    \begin{subfigure}[b]{0.4\textwidth}
        \centering
        \includegraphics[width=\linewidth]{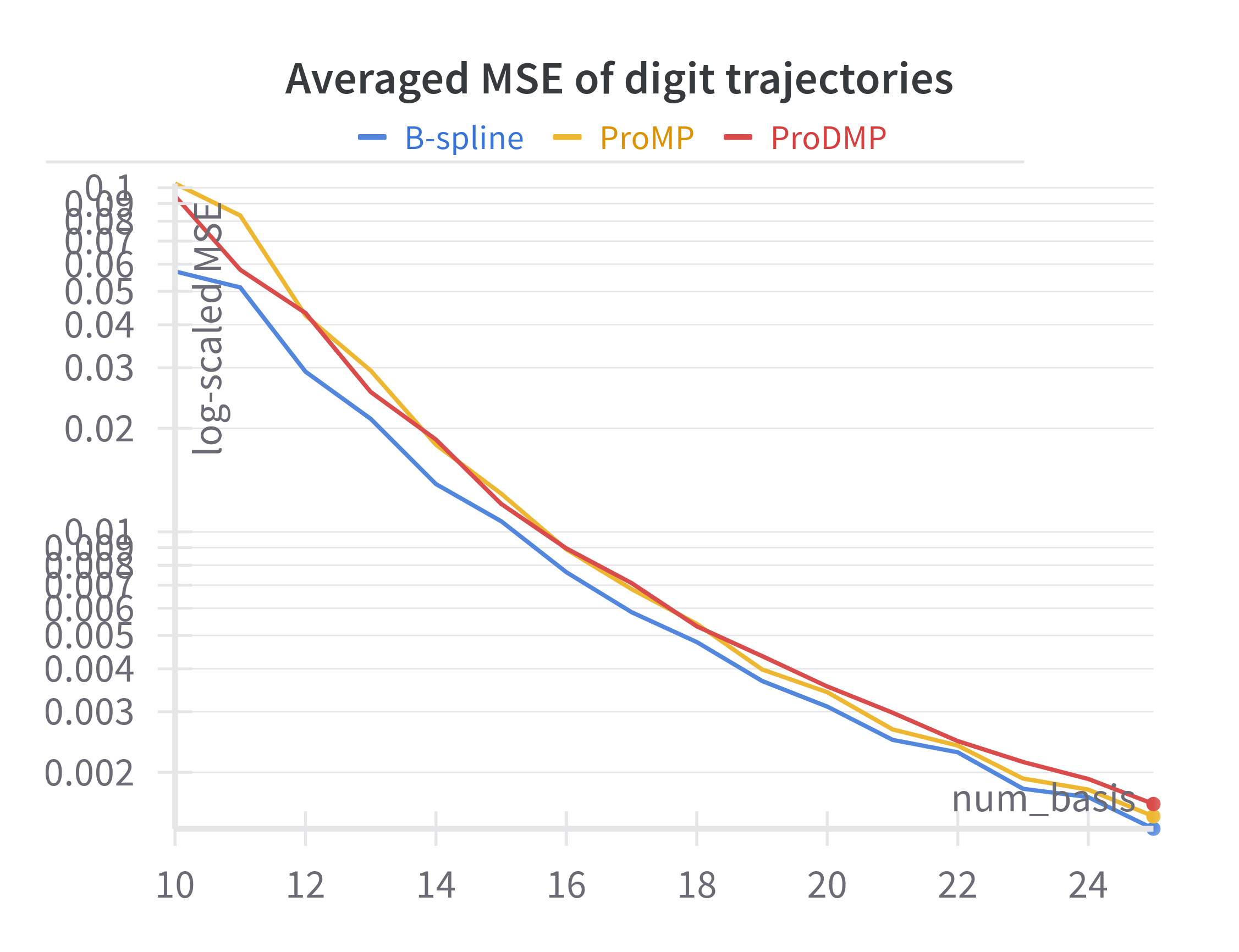}
        % \caption{The log-scaled averaged regression MSE loss on 20000 3-second digit-writing trajectories, by applying B-spline and MPs with different numbers of basis functions.}
        \caption{}
        \label{fig:regloss}
    \end{subfigure}
    \hfill
    \begin{subfigure}[b]{0.4\textwidth}
        \centering
        \includegraphics[width=\linewidth]{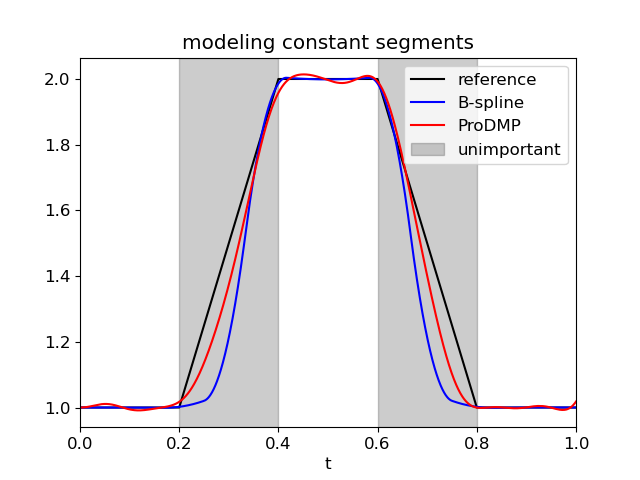}
        % \caption{regressing B-spline and ProDMP on a trajectory with three constant segments, where the grey shadowed transition segments are unimportant and not considered in regression. ProDMP exhibits obvious wiggles in the all constant segments $[0,\  0.2],\ [0.4,\ 0.6],\ [0.8,\ 1]$}
        \caption{}
        \label{fig:consseg}
    \end{subfigure}
    \caption{(a) The log-scaled averaged regression MSE loss on 20000 3-second digit-writing trajectories, by applying B-spline and MPs with different numbers of basis functions. (b) Regressing B-spline and ProDMP on a trajectory with three constant segments, where the grey-shadowed transition segments are unimportant and not considered in regression. ProDMP exhibits obvious wiggles in the all constant segments $[0,\  0.2],\ [0.4,\ 0.6],\ [0.8,\ 1]$}
    % \label{fig:main_label}
\end{figure}

% B-spline advantages in regression and planning joy tasks
We first show that B-splines are generally more expressive than ProDMP and ProMP. As illustrated in \ref{fig:regloss}, by regressing B-splines and MPs on a set of digit trajectories, B-splines achieve a lower mean squared error (MSE) loss.
Additionally, an effective trajectory representation should be capable of representing constant segments, corresponding to a robot remaining at specific configurations. For a trajectory representation that is a linear combination of basis functions, this requirement is satisfied by normalized basis functions, as is the case for both B-spline and ProMP. However, as depicted in \ref{fig:consseg}, ProDMP is unable to represent constant segments.

\begin{figure}[t!]
    \centering
    \begin{subfigure}[b]{0.32\textwidth}
        \centering
        \includegraphics[width=\linewidth]{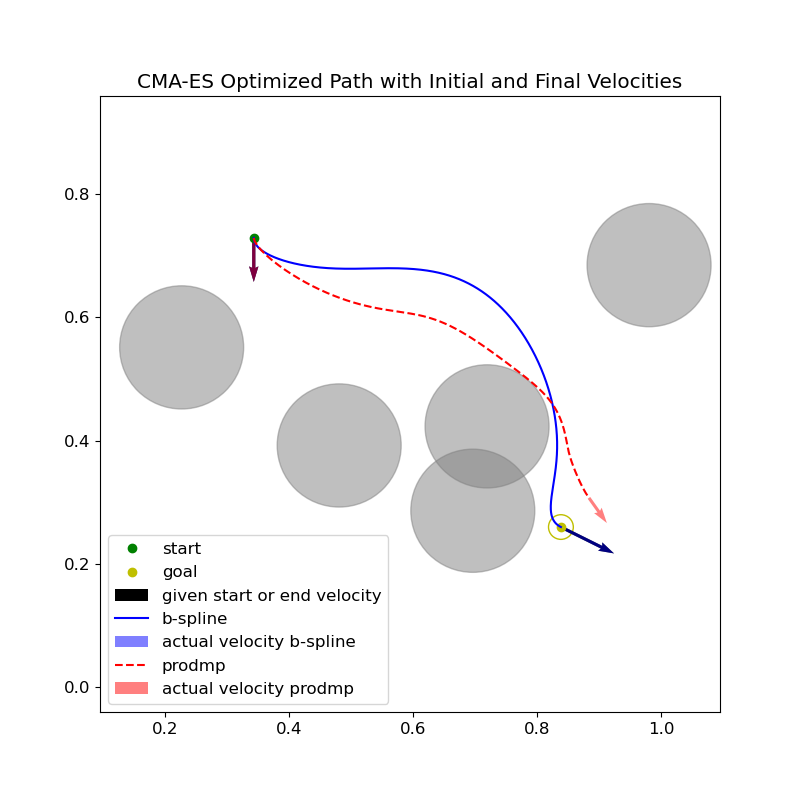}
        \caption{}
        \label{fig:ppath}
    \end{subfigure}
    \hfill
    \begin{subfigure}[b]{0.32\textwidth}
        \centering
        \includegraphics[width=\linewidth]{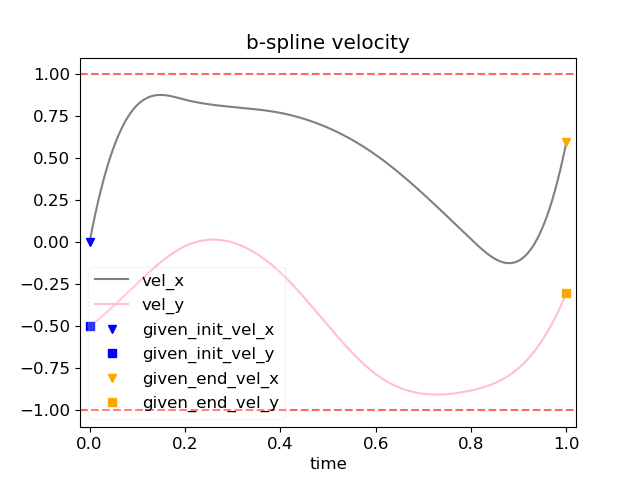}
        \caption{}
        \label{fig:pbsp_vel}
    \end{subfigure}
    \hfill
    \begin{subfigure}[b]{0.32\textwidth}
        \centering
        \includegraphics[width=\linewidth]{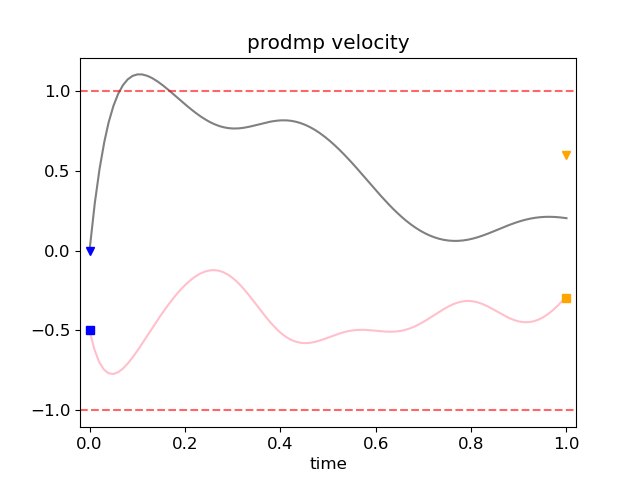}
        \caption{}
        \label{fig:pprod_vel}
    \end{subfigure}
    \hfill
    \caption{Using CMA-ES \cite{hansen_cma_2023} algorithm for 50 iterations to generate trajectories from given initial states to reach goal states while avoiding obstacles. (a) Generated best trajectory path within 50 iterations. (b),(c) The velocity profile of the B-spline and ProDMP trajectory, where the red dash lines are the velocity bounds.}
    \label{fig:planning}
\end{figure}

B-spline offers significant advantages as a trajectory generator for planning and RL tasks, particularly in their ability to specify arbitrary boundary conditions. In contrast, ProDMP guarantees only the satisfaction of initial position and velocity, while ProMP does not support the specification of either initial or final conditions.
As demonstrated in \ref{fig:planning}, a goal-reaching toy task, where a trajectory must be generated from given initial conditions to reach a target with a desired velocity while avoiding obstacles. In addition, the trajectory should satisfy a given velocity bound. By setting boundary conditions, B-splines complete the task within a few iterations, where it only learns to avoid obstacles. Besides, utilizing the convex hull property of the B-spline, we can simply set soft constraints for velocity control points to satisfy velocity bounds. In contrast, accommodating above requirements through reward shaping in ProDMP is untrivial.

% As shown in \ref{fig:joint_vel}, the B-spline ensures satisfying the prescribed end velocity, but ProDMP can not.

\subsection{BMP for Imitation Learning and Reinforcement Learning}
% B-spline within ProMP framework, BMP for IL and RL

% \begin{figure}[ht!]
%     \centering
%     \includegraphics[width=0.8\textwidth]{figures/ENNN_block.png}
%     \caption{The encoder-decoder neural network architecture from \citep{li_prodmp_2023}}
%     \label{fig:nn_block}
% \end{figure}

\textbf{Imitation Learning} We embed BMP into a neural network as a generative model, to generate digit-writing trajectories conditioned on digit images from the synthetic-MNIST dataset \citep{PAHIC2020121}. We adopt an encoder-decoder neural network architecture and its training procedure proposed in \citep{li_prodmp_2023}, where the neural network takes a set of digit writing images as input and predict the mean $\bm{\mu_{w}}$ and the Cholesky decomposition $\bm{L_{w}}$ of the covariance $\bm{\Sigma_{w}}$ of the weights distribution of the BMP. Then BMP maps this weights distribution to the trajectory distribution $\mathcal{N}(\bm{\mu_\Lambda,\Sigma_\Lambda)}$.
%这段放到appendix
% This architecture as shown in \Cref{fig:nn_block}, consists of four blocks. The Encoders networks $E_\bm{\mu}$ and $E_{\bm{\sigma}^2}$ compute a latent representation $\bm{r}_m$ and an uncertainty $\bm{\sigma}^2_{\bm{r}_m}$ for each observation $\bm{o}_m \in \mathcal{O}$. The Bayesian aggregator $\bm{A}$ as a functional aggregates the latent observations ${\bm{r}_m}$ weighted by inverses of their uncertainties ${\bm{\sigma}^2_{\bm{r}_m}}$, into a latent state posterior $p(\bm{z}|\bm{O})$, which is a factorized multivariate Gaussian distribution $\mathcal{N}(\bm{z}|\bm{\mu_z}, \bm{\sigma_z^2})$.
% The Decoder networks $D_{\bm{\mu}}$, $D_{\bm{L}}$ compute the mean $\bm{\mu_{w}}$ and the Cholesky decomposition $\bm{L_{w}}$ of the covariance $\bm{\Sigma_{w}}$ of the weights distribution based on latent variable posterior. Finally, the MP layer maps this weights distribution to the trajectory distribution $\mathcal{N}(\bm{\mu_\Lambda,\Sigma_\Lambda)}$.
%放到appendix

\begin{figure}[h!]
    \centering

    \begin{subfigure}{\textwidth} % First subfigure block
        \centering
        \includegraphics[width=\textwidth]{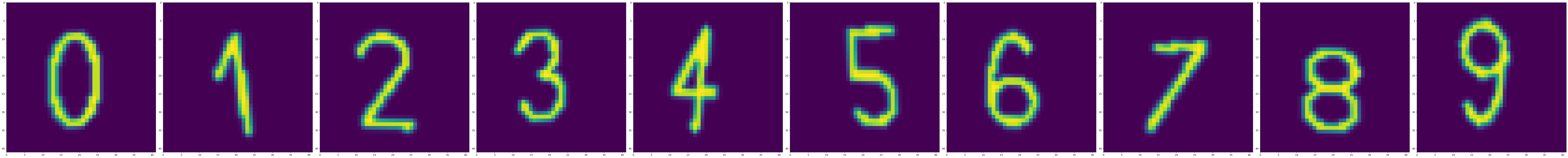} \\
        \includegraphics[width=\textwidth]{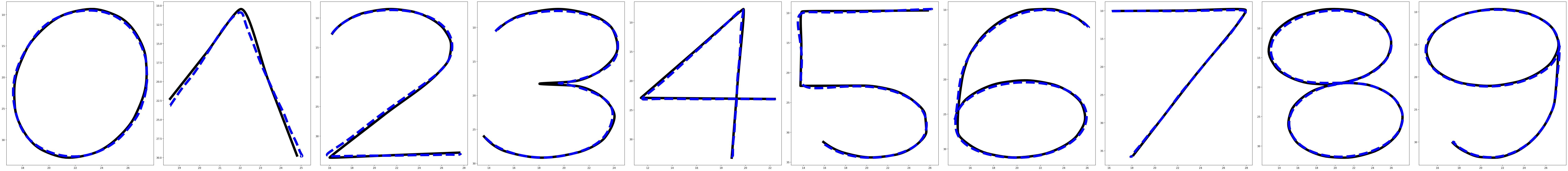}
        \caption{}
        % \caption{Digits images and reconstructed trajectories}
        \label{fig:digits_rec}
    \end{subfigure}

    \vspace{1em} % Add vertical space between rows of subfigures

    \begin{subfigure}{\textwidth} % Second subfigure block (left)        
        \centering
        \includegraphics[width=0.2\textwidth]{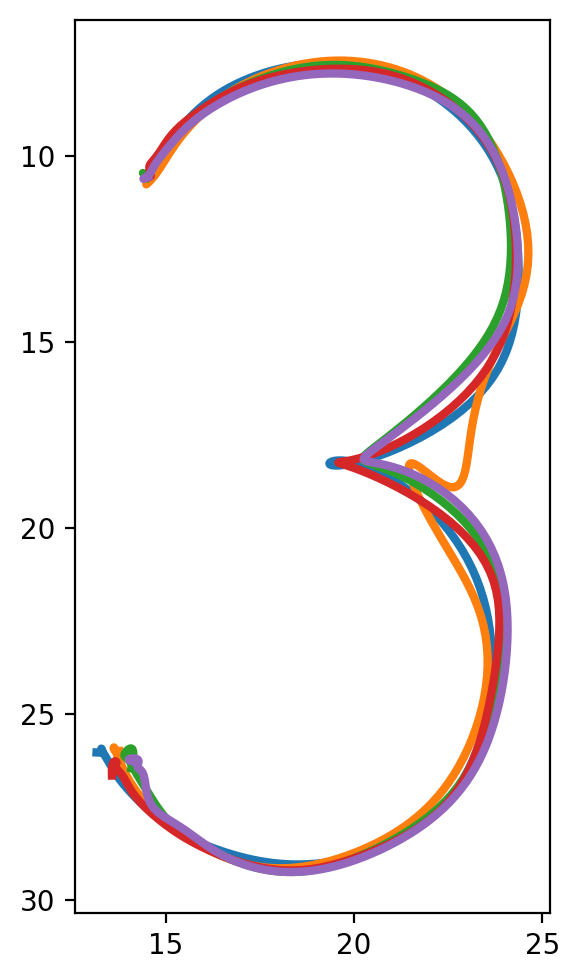}
        \includegraphics[width=0.2\textwidth]{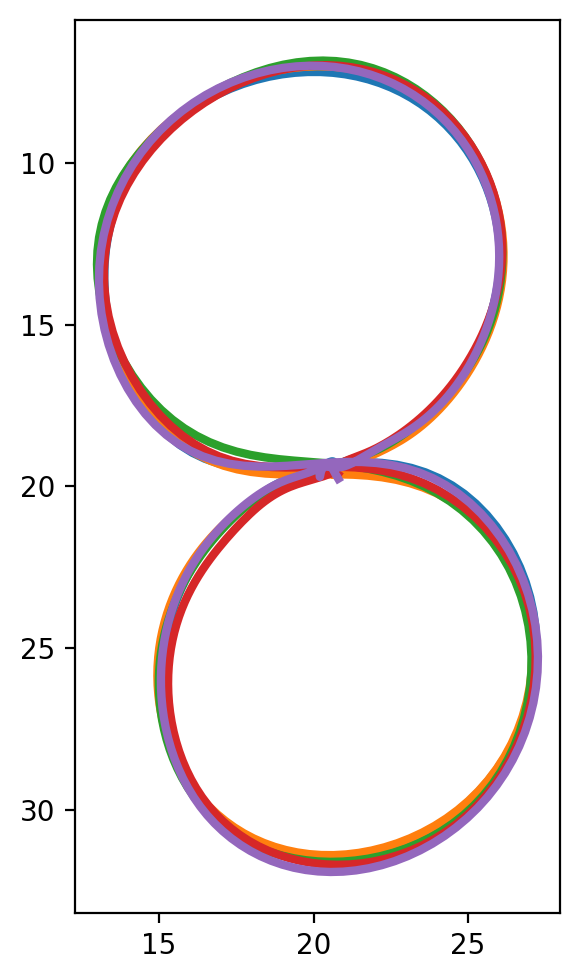}
        \caption{}
    \end{subfigure}

    \caption{(a) A batch of digit images and reconstructed trajectories using BMP. (b) Sampling trajectories from predicted distribution conditioned on an image of digit '3' and '8'.}
    \label{fig:digitsimi}
\end{figure}

Since we maintain a distribution of trajectories through BMP, we can leverage probabilistic modeling techniques to train this generative model. Specifically, we aim to minimize the negative log-likelihood of the ground truth trajectories on the predicted trajectory distribution, i.e., \(- \log \mathcal{N}(\bm{\Lambda|\mu_\Lambda, \Sigma_\Lambda)}\). Here, \(\bm{\Lambda} = \{\bm{y}_t\}_{t=0..T}\) is the trajectory ground truth, and \(\bm{\mu_\Lambda, \Sigma_\Lambda}\) the mean and covariance of the predicted trajectory distribution conditioned on a set of image observations \(\mathcal{O}\). However, directly computing the log-likelihood of that a high-dimensional Gaussian distribution during each training step is computationally infeasible due to the need to invert the \(TD \times TD \) dimension covariance matrix \( \bm{\Sigma_\Lambda}\). To get around this problem while keep capturing the temporal correlations in trajectories, \( J\) time point pairs \(\{(t, t')_j\}_{j=1,...,J}\) are randomly sampled and for each time point pair a log-likelihood is calculated, where only \( J\) times \(2D \times 2D \) matrix inversion is performed, resulting in an alternative loss function
\begin{equation}
    \mathcal{L}_{\bm{\theta}}(\bm{\Lambda} \mid \mathcal{O}) = -\frac{1}{J} \sum_{j=1}^{J} \log \mathcal{N}\left( \bm{y}_{(t,t')_j} \mid \bm{\mu}_{(t,t')_j}, \bm{\Sigma}_{(t,t')_j} \right). \label{eq:imiloss}
\end{equation}
Although this training objective is biased from the exact negative log-likelihood of trajectories on the predicted correlated trajectory distribution, it captures correlations and reconstructs high-quality trajectories as shown in \ref{fig:digitsimi}. Using BMP and this training procedure allows training a generative model to sample trajectories.

% \begin{figure}[h!]
%     \centering

%     \begin{subfigure}{\textwidth} % First subfigure block
%         \centering
%         \includegraphics[width=\textwidth]{figures/digit_image.png} \\
%         \includegraphics[width=\textwidth]{figures/recon.png}
%         \caption{Digits images and reconstructed trajectories}
%         \label{fig:digits_rec}
%     \end{subfigure}

%     \vspace{1em} % Add vertical space between rows of subfigures

%     \begin{subfigure}{0.35\textwidth} % Second subfigure block (left)        
%         \centering
%         \includegraphics[width=\textwidth]{figures/3.png}
%         \caption{}
%     \end{subfigure}
%     \hfill
%     \begin{subfigure}{0.35\textwidth} % Second subfigure block (right)
%         \centering
%         \includegraphics[width=\textwidth]{figures/8.png}
%         \caption{}
%     \end{subfigure}

%     \caption{(a) A batch of digit images and reconstructed trajectories using B-spline trajectory. (b) Sampling trajectories from predicted distribution conditioned on image of digit '3' and '8'.}
%     \label{fig:digitsimi}
% \end{figure}

% \begin{figure}[ht!]
%     \centering
%     \includegraphics[width=0.8\linewidth]{figures/Box push success rate.png}
%     \caption{The averaged success rate of 4 random seeds. Both B-spline and ProDMP uses in total 9 basis functions. The basis functions of ProDMP are scaled into same value range to make it work.}
%     \label{fig:boxsucc}
% \end{figure}
\begin{figure}[ht!]
    \centering
    \begin{subfigure}[b]{0.6\textwidth}
        \centering
        \includegraphics[width=\linewidth]{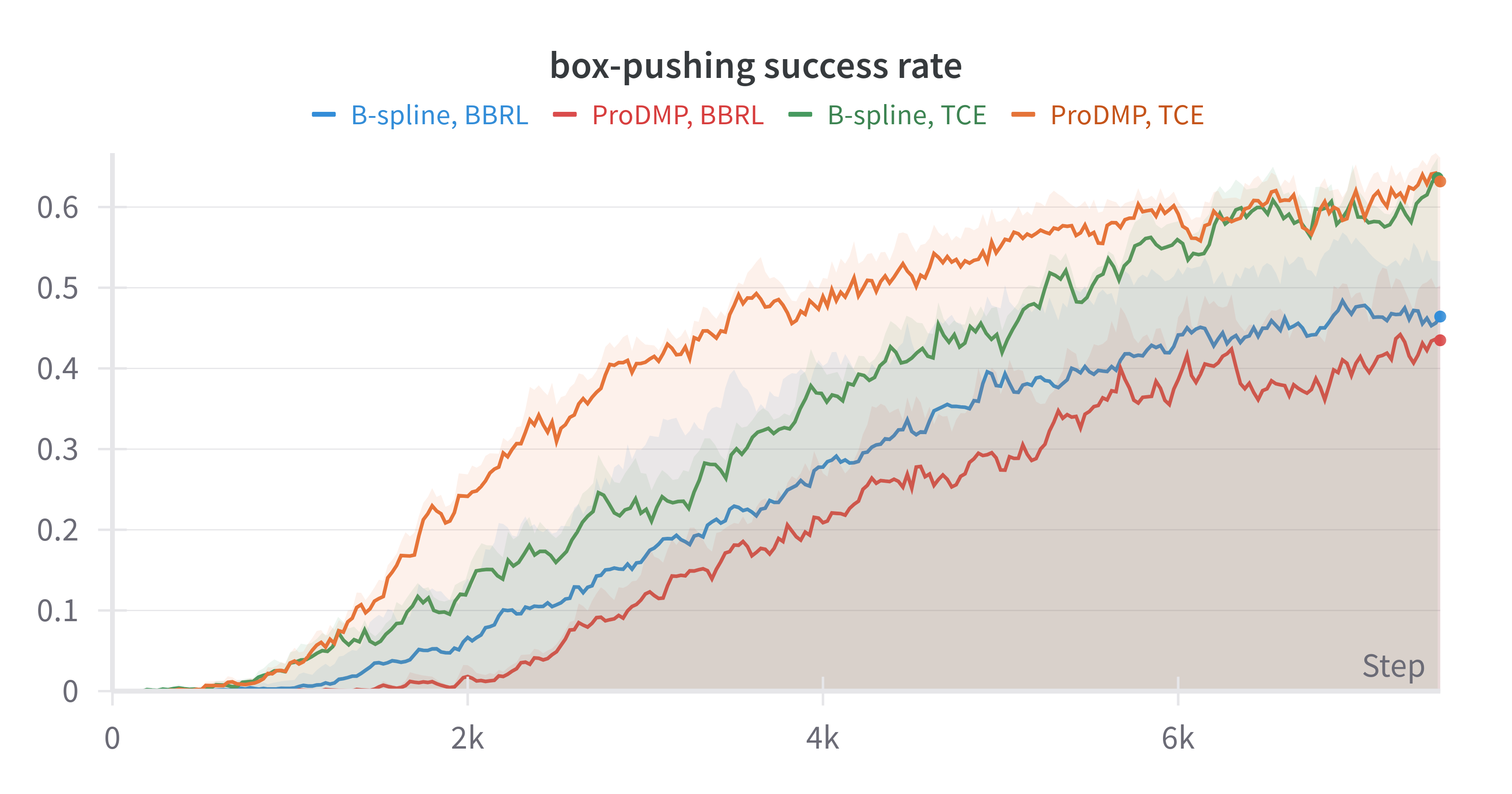}
        % \caption{The averaged success rate of 4 random seeds. Both B-spline and ProDMP use in total 9 basis functions. The basis functions of ProDMP are scaled into the same value range to make it work.}
        \caption{}
        \label{fig:boxsucc}
    \end{subfigure}
    \hfill
    \begin{subfigure}[b]{0.35\textwidth}
        \centering
        \includegraphics[width=\linewidth]{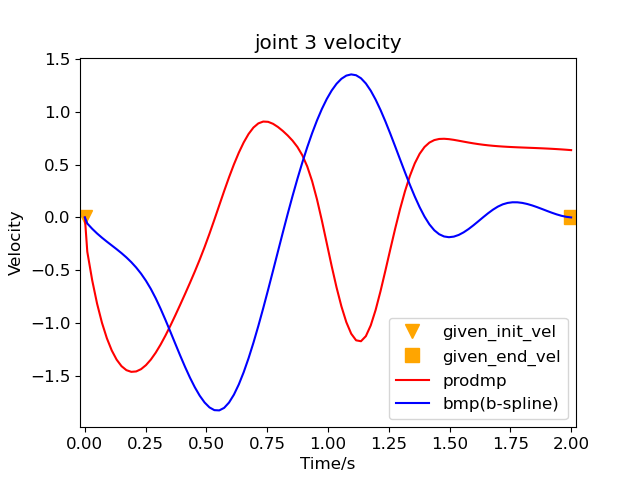}
        % \caption{The joint velocity of joint 3, where the velocity at 0s and 2s should be 0. The velocity profile from 0.01s to 2s of BMP (B-spline) and ProDMP are plotted.}
        \caption{}
        \label{fig:joint_vel}
    \end{subfigure}
    \caption{(a) The averaged success rate of 4 random seeds. Both BMP (B-spline) and ProDMP use in total 9 basis functions. The basis functions of ProDMP are scaled into the same value range to make it work. (b) The joint velocity profile of joint 3, where the velocity at 0s and 2s should be 0.}
    \label{fig:rl}
\end{figure}

% \subsection{Reinforcemt Learning}
\textbf{Episodic Reinforcement Learning (ERL)} treats RL problem as a contextual black-box optimization problem, where the action sequence for an episode is parameterized by $\bm{\omega}$ and the whole policy roll-out process is considered as one entity, a black-box. It aims to maximize the expected return of an episode policy roll-out $R(\bm{w,c})$ by optimizing the contextual search distribution $\pi(\bm{\omega}|\bm{c})$ of the parameters $\omega$, i.d.
\begin{equation}
    \arg\max_{\pi(\bm{w|c})} \mathbb{E}_{p(c)} \left[ \mathbb{E}_{\pi(\bm{w|c})} [R(\bm{w,c})] \right], \label{eq:erlj}
\end{equation}
where $\bm{c}$ is the context and $p(\bm{c})$ is the distribution of the context. Compared to step-based RL, the underlying world model is not necessarily a Markov Decision Process (MDP). Combining with MP, ERL can generate smoother trajectories and allow consistent oriented exploration.
We adopt two ERL algorithms:

\textbf{Deep-Black Box Reinforcement Learning (BBRL)} \citep{Otto_bbrl_2022}, whose surrogate learning objective is
\begin{equation}
    \hat{J}(\pi_{\bm{\theta}}, \pi_{\bm{\theta}_{\text{old}}}) = \mathbb{E}_{(\bm{c, w}) \sim p(\bm{c}), \pi_{\bm{\theta}_{\text{old}}}} \left[ \frac{\pi_{\bm{\theta}}(\bm{w | c})}{\pi_{\bm{\theta}_{\text{old}}}(\bm{w | c})} A^{\pi_{\bm{\theta}_{\text{old}}}}(\bm{c, w}) \right], \label{eq:bbrlj}
\end{equation}
where $A^{\pi_{\bm{\theta}_{\text{old}}}}(\bm{c, w})$ is an advantage function. The objective is to maximize the advantage-weighted likelihood at the parameter level. BBRL can be simply applied to the B-spline.

\textbf{Temporal Correlated Exploration (TCE)} \citep{Li_tce_2024} divides the whole trajectory into $K$ segments $\left[ \bm{y}_t \right]_{t=t_k:t'_k}$. By maintaining trajectory distribution and taking the same philosophy in \eqref{eq:imiloss}, the learning objective takes the form
\begin{equation}
    J(\bm{\theta}) = \mathbb{E}_{\pi_{\text{old}}} \left[ \frac{1}{K} \sum_{k=1}^{K} \frac{p_{\pi_{\text{new}}} \left( [\mathbf{y}_t]_{t=t_k:t'_k} \middle| \mathbf{s} \right)}{p_{\pi_{\text{old}}} \left( [\mathbf{y}_t]_{t=t_k:t'_k} \middle| \mathbf{s} \right)} A^{\pi_{\text{old}}} \left( s_{t_k}, [\mathbf{y}_t]_{t=t_k:t'_k} \right) \right].
\end{equation}
 In contrast to BBRL, TCE maximizes the advantage-weighted likelihood for each segment at the trajectory level, which allows a finer-grained update of the policy while using the same roll-out data as in BBRL. This training scheme requires trajectory distribution, thus can only apply to BMP.

We demonstrate using BMP within both algorithms in a contact-rich manipulation task, the box-pushing task, where the objective is to move a box to a specified goal position with a specified goal orientation using a rod fixed on the end effector of a seven DoF Franka Emika Panda robot. 
The result is evaluated in terms of success rate, as given in \ref{fig:boxsucc}. With the BBRL algorithms, the B-spline outperforms the ProDMP, possibly because the locally supported basis function benefits the optimization process. By maintaining trajectory distribution and using TCE, the convergence rate and final success rate improve significantly. The ProDMP shows faster growth in TCE at the early stage, we hypothesize that this is because ProDMP has a goal basis spanning the whole trajectory duration, which can enforce a temporal correlation in the trajectory distribution at the early stage. This may benefit the TCE training scheme. Generally, BMP shows comparable performance to ProDMP and ensures satisfying prescribed end conditions, but ProDMP is unable to ensure end velocity conditions, as shown in \ref{fig:joint_vel}.

In summary, B-spline is generally a better trajectory representation than previous MPs. Utilizing B-spline as a probabilistic MP, the BMP, enables the probabilistic modeling for B-spline and enhances its application in IL and RL.

%===============================================================================

% \section{Experimental Results}
% \label{sec:result}

%===============================================================================

\section{Conclusion}
\label{sec:conclusion}
Our work presents using B-spline as MP in the ProMP framework, the BMP. This allows using probabilistic techniques for learning and ensuring satisfying boundary conditions. BMP can be a unified MP framework for imitation learning and reinforcement learning.
In future work, we will further investigate the possible advantages of BMP. Especially, we are investigating how to efficiently and effectively utilize the convex hull property of B-spline to impose kinodynamic constraints in BBRL and TCE algorithms for practical reinforcement learning tasks.

%===============================================================================

\clearpage
% The acknowledgments are automatically included only in the final and preprint versions of the paper.
\acknowledgments{If a paper is accepted, the final camera-ready version will (and probably should) include acknowledgments. All acknowledgments go at the end of the paper, including thanks to reviewers who gave useful comments, to colleagues who contributed to the ideas, and to funding agencies and corporate sponsors that provided financial support.}

%===============================================================================

\bibliography{example}  % .bib

\begin{thebibliography}{16}
\providecommand{\natexlab}[1]{#1}
\providecommand{\url}[1]{\texttt{#1}}
\expandafter\ifx\csname urlstyle\endcsname\relax
  \providecommand{\doi}[1]{doi: #1}\else
  \providecommand{\doi}{doi: \begingroup \urlstyle{rm}\Url}\fi

\bibitem[Ijspeert et~al.(2013)Ijspeert, Nakanishi, Hoffmann, Pastor, and Schaal]{Ijspeert_2013}
A.~J. Ijspeert, J.~Nakanishi, H.~Hoffmann, P.~Pastor, and S.~Schaal.
\newblock Dynamical movement primitives: Learning attractor models for motor behaviors.
\newblock \emph{Neural Computation}, 25\penalty0 (2):\penalty0 328--373, 2013.
\newblock \doi{10.1162/NECO_a_00393}.

\bibitem[Paraschos et~al.(2013)Paraschos, Daniel, Peters, and Neumann]{NIPS2013_e53a0a29}
A.~Paraschos, C.~Daniel, J.~R. Peters, and G.~Neumann.
\newblock Probabilistic movement primitives.
\newblock In C.~Burges, L.~Bottou, M.~Welling, Z.~Ghahramani, and K.~Weinberger, editors, \emph{Advances in Neural Information Processing Systems}, volume~26. Curran Associates, Inc., 2013.
\newblock URL \url{https://proceedings.neurips.cc/paper_files/paper/2013/file/e53a0a2978c28872a4505bdb51db06dc-Paper.pdf}.

\bibitem[Zhou et~al.(2019)Zhou, Gao, and Asfour]{zhou2019learning}
Y.~Zhou, J.~Gao, and T.~Asfour.
\newblock Learning via-point movement primitives with inter-and extrapolation capabilities.
\newblock In \emph{2019 IEEE/RSJ International Conference on Intelligent Robots and Systems (IROS)}, pages 4301--4308. IEEE, 2019.

\bibitem[Li et~al.(2023)Li, Jin, Volpp, Otto, Lioutikov, and Neumann]{li_prodmp_2023}
G.~Li, Z.~Jin, M.~Volpp, F.~Otto, R.~Lioutikov, and G.~Neumann.
\newblock Prodmp: A unified perspective on dynamic and probabilistic movement primitives.
\newblock \emph{IEEE Robotics and automation letters}, page 1–8, 2023.
\newblock ISSN 2377-3766.
\newblock \doi{10.1109/LRA.2023.3248443}.

\bibitem[Usenko et~al.(2017)Usenko, Von~Stumberg, Pangercic, and Cremers]{usenko_real-time_2017}
V.~Usenko, L.~Von~Stumberg, A.~Pangercic, and D.~Cremers.
\newblock Real-time trajectory replanning for {MAVs} using uniform {B}-splines and a {3D} circular buffer.
\newblock In \emph{2017 {IEEE}/{RSJ} {International} {Conference} on {Intelligent} {Robots} and {Systems} ({IROS})}, pages 215--222, Vancouver, BC, Sept. 2017. IEEE.
\newblock ISBN 978-1-5386-2682-5.
\newblock \doi{10.1109/IROS.2017.8202160}.
\newblock URL \url{http://ieeexplore.ieee.org/document/8202160/}.

\bibitem[Zhou et~al.(2019)Zhou, Gao, Wang, Liu, and Shen]{zhou_robust_2019}
B.~Zhou, F.~Gao, L.~Wang, C.~Liu, and S.~Shen.
\newblock Robust and {Efficient} {Quadrotor} {Trajectory} {Generation} for {Fast} {Autonomous} {Flight}.
\newblock \emph{IEEE Robotics and Automation Letters}, 4\penalty0 (4):\penalty0 3529--3536, Oct. 2019.
\newblock ISSN 2377-3766.
\newblock \doi{10.1109/LRA.2019.2927938}.
\newblock URL \url{https://ieeexplore.ieee.org/document/8758904}.
\newblock Conference Name: IEEE Robotics and Automation Letters.

\bibitem[Kicki et~al.(2023)Kicki, Liu, Tateo, Bou-Ammar, Walas, Skrzypczyński, and Peters]{kicki_fast_2023}
P.~Kicki, P.~Liu, D.~Tateo, H.~Bou-Ammar, K.~Walas, P.~Skrzypczyński, and J.~Peters.
\newblock Fast {Kinodynamic} {Planning} on the {Constraint} {Manifold} with {Deep} {Neural} {Networks}, Jan. 2023.
\newblock URL \url{http://arxiv.org/abs/2301.04330}.
\newblock arXiv:2301.04330 [cs].

\bibitem[Kicki et~al.(2024)Kicki, Tateo, Liu, Guenster, Peters, and Walas]{kicki2024bridginggaplearningtoplanmotion}
P.~Kicki, D.~Tateo, P.~Liu, J.~Guenster, J.~Peters, and K.~Walas.
\newblock Bridging the gap between learning-to-plan, motion primitives and safe reinforcement learning, 2024.
\newblock URL \url{https://arxiv.org/abs/2408.14063}.

\bibitem[Li et~al.(2024)Li, Zhou, Roth, Thilges, Otto, Lioutikov, and Neumann]{Li_tce_2024}
G.~Li, H.~Zhou, D.~Roth, S.~Thilges, F.~Otto, R.~Lioutikov, and G.~Neumann.
\newblock Open the black box: Step-based policy updates for temporally-correlated episodic reinforcement learning.
\newblock In \emph{ICLR 2024 : The Twelfth International Conference on Learning Representations, Vienna, 7th-11th May 2024}. {International Conference on Learning Representations, ICLR}, 2024.

\bibitem[Otto et~al.(2022)Otto, Celik, Zhou, Ziesche, Ngo, and Neumann]{Otto_bbrl_2022}
F.~Otto, O.~Celik, H.~Zhou, H.~Ziesche, V.~A. Ngo, and G.~Neumann.
\newblock Deep black-box reinforcement learning with movement primitives.
\newblock In \emph{6th Annual Conference on Robot Learning (CoRL 2022)}, volume 205 of \emph{Proceedings of Machine Learning Research}, pages 1244--1265. {Machine Learning Research Press (ML Research Press)}, 2022.

\bibitem[Otto et~al.(2023)Otto, Zhou, Celik, Li, Lioutikov, and Neumann]{otto2023mp3}
F.~Otto, H.~Zhou, O.~Celik, G.~Li, R.~Lioutikov, and G.~Neumann.
\newblock Mp3: Movement primitive-based (re-) planning policy.
\newblock \emph{arXiv preprint arXiv:2306.12729}, 2023.

\bibitem[Li et~al.(2024)Li, Tian, Zhou, Jiang, Lioutikov, and Neumann]{li2024top}
G.~Li, D.~Tian, H.~Zhou, X.~Jiang, R.~Lioutikov, and G.~Neumann.
\newblock Top-erl: Transformer-based off-policy episodic reinforcement learning.
\newblock \emph{arXiv preprint arXiv:2410.09536}, 2024.

\bibitem[Saveriano et~al.(2023)Saveriano, Abu-Dakka, Kramberger, and Peternel]{saveriano_dynamic_2021}
M.~Saveriano, F.~J. Abu-Dakka, A.~Kramberger, and L.~Peternel.
\newblock Dynamic movement primitives in robotics: A tutorial survey.
\newblock \emph{The International Journal of Robotics Research}, 42\penalty0 (13):\penalty0 1133--1184, 2023.
\newblock \doi{10.1177/02783649231201196}.
\newblock URL \url{https://doi.org/10.1177/02783649231201196}.

\bibitem[Prautzsch et~al.(2002)Prautzsch, Boehm, and Paluszny]{prautzsch2002bezier}
H.~Prautzsch, W.~Boehm, and M.~Paluszny.
\newblock \emph{B{\'e}zier and B-Spline Techniques}.
\newblock Springer, Berlin, Heidelberg, 2002.
\newblock ISBN 978-3-642-05240-8.
\newblock \doi{10.1007/978-3-662-04947-3}.

\bibitem[Hansen(2023)]{hansen_cma_2023}
N.~Hansen.
\newblock The {CMA} {Evolution} {Strategy}: {A} {Tutorial}, Mar. 2023.
\newblock URL \url{http://arxiv.org/abs/1604.00772}.
\newblock arXiv:1604.00772 [cs, stat].

\bibitem[Pahič et~al.(2020)Pahič, Ridge, Gams, Morimoto, and Ude]{PAHIC2020121}
R.~Pahič, B.~Ridge, A.~Gams, J.~Morimoto, and A.~Ude.
\newblock Training of deep neural networks for the generation of dynamic movement primitives.
\newblock \emph{Neural Networks}, 127:\penalty0 121--131, 2020.
\newblock ISSN 0893-6080.
\newblock \doi{https://doi.org/10.1016/j.neunet.2020.04.010}.
\newblock URL \url{https://www.sciencedirect.com/science/article/pii/S0893608020301301}.

\end{thebibliography}

\appendix

\end{document}